\documentclass[10pt,twocolumn,letterpaper]{article}

\usepackage{cvpr}
\usepackage{times}
\usepackage{epsfig}
\usepackage{graphicx}
\usepackage{amsmath}
\usepackage{amssymb}
\usepackage{adjustbox}
\usepackage{pifont}

\usepackage[pagebackref=true,breaklinks=true,letterpaper=true,colorlinks,bookmarks=false]{hyperref}
\newcommand{\cmark}{\ding{51}}%
\newcommand{\xmark}{\ding{55}}%
\cvprfinalcopy 

\def\cvprPaperID{8466} 

\ifcvprfinal\fi
\begin{document}
	
	\title{Plug \& Play Convolutional Regression Tracker \\ for Video Object Detection}
	
	\author{Ye Lyu\\
		University of Twente
		\and
		Michael Ying Yang\\
		University of Twente
		\and
		George Vosselman\\
		University of Twente
		\and
		Gui-Song Xia\\
		Wuhan University
	}
	
	\maketitle
	
	\begin{abstract}
		Video object detection targets to simultaneously localize the bounding boxes of the objects and identify their classes in a given video.
		One challenge for video object detection is to consistently detect all objects across the whole video. As the appearance of objects may deteriorate in some frames, features or detections from the other frames are commonly used to enhance the prediction. In this paper, we propose a Plug \& Play scale-adaptive convolutional regression tracker for the video object detection task, which could be easily and compatibly implanted into the current state-of-the-art detection networks. As the tracker reuses the features from the detector, it is a very light-weighted increment to the detection network. 
		The whole network performs at the speed close to a standard object detector. With our new video object detection pipeline design, image object detectors can be easily turned into efficient video object detectors without modifying any parameters. The performance is evaluated on the large-scale ImageNet VID dataset. Our Plug \& Play design improves mAP score for the image detector by around 5\% with only little speed drop.
	\end{abstract}
	
	\section{Introduction}
	Last several years have witnessed the fast development of the computer vision in scene understanding, especially the fundamental object detection task. Object detection is to simultaneously localize the bounding boxes of the objects and classify their class types in an image. Video object detection extends the task to videos. It requires the detector to utilize multiple frames in a video to provide consistent predictions over time, which is another fundamental task for computer vision. The large scale datasets and the convolutional neural networks have been two of the main propellers for such visual understanding tasks. Both of the tasks have received much attention since the introduction of ImageNet object detection (DET) challenge and ImageNet video object detection (VID) challenge in ILSVRC 2015~\cite{imagenet}. The video object detection is more challenging than the image object detection due to that the appearance of the objects may deteriorate significantly in some frames caused by motion blur, video defocus, part occlusion or rare poses. However, the rich context information in temporal domain provides the clues and the opportunity to classify the objects.
	
	\begin{figure}
		\begin{center}
			\includegraphics[width=1.0\linewidth]{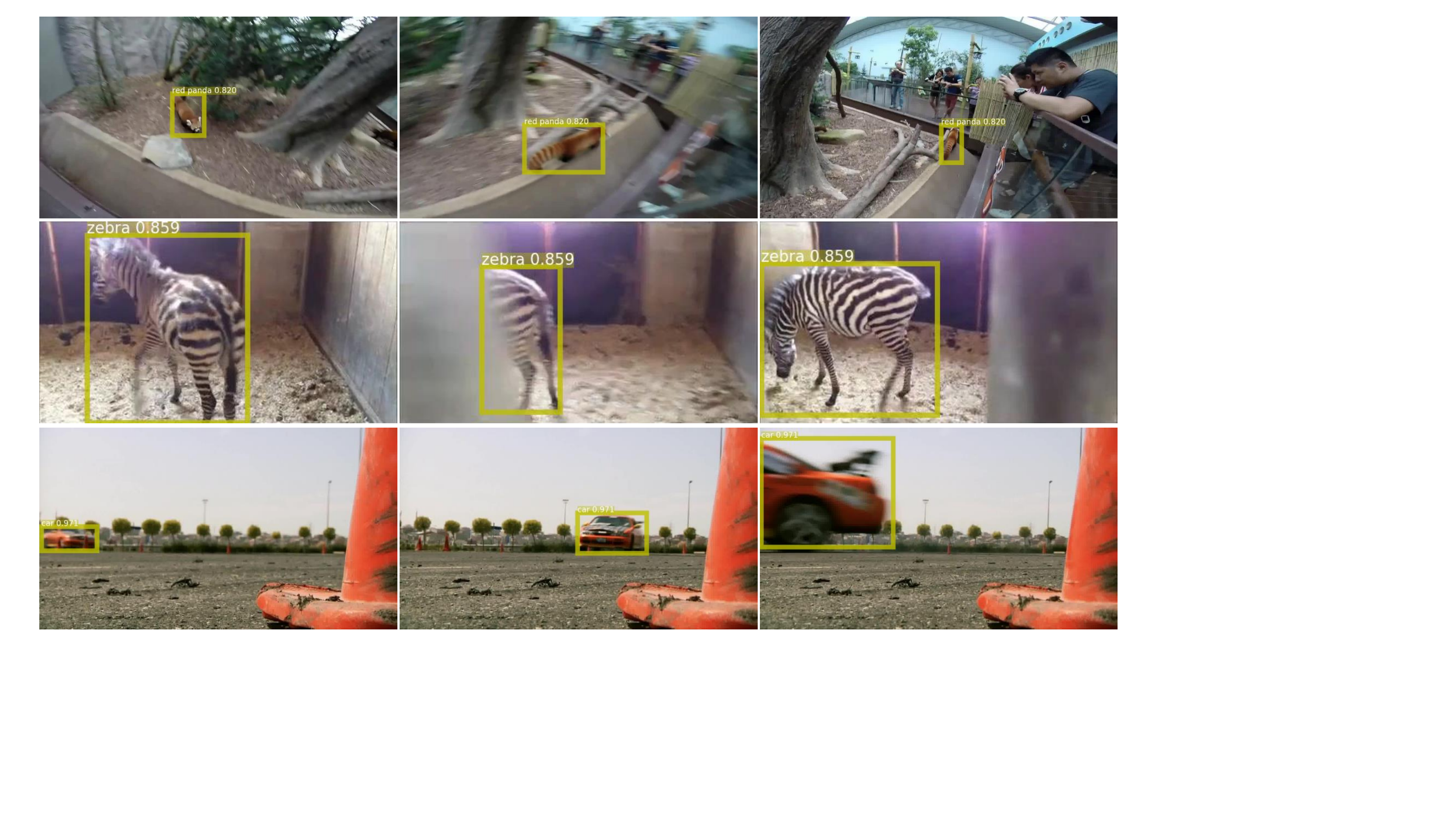}
		\end{center}
		\caption{Examples of object detection in videos by our method. The scores are consistent over the long term even if the image quality decays or the objects are partially occluded or in rare poses.}
		\label{fig:vod_eg_fg1}
	\end{figure}
	
	One basic video object detection method is to detect objects from individual image first and further apply a post-processing to link and re-score the object detection results in the video~\cite{seqnms,TCNN}. The linking can be based on either the appearance similarity scores~\cite{track_cond} or the external video object tracker~\cite{TCNN}. The problem for such methods is that they require good enough initial image object detection results. If the initial detection is lost or the localization of the initial detection is not accurate, the post-processing performance may not be ideal. Feature aggregation is another widely used method for video object detection. By using features from multiple frames, a more temporally coherent and robust detection result can be achieved. The problem for short term feature aggregation~\cite{stmn,stsn,FGFA,manet} is that they pre-define a limited range of frames for the detection in each frame and the long range coherency cannot be preserved and utilized. Long term feature aggregation achieves better results ~\cite{vod_selsa,vod_relation_distillation,vod_lrtr}. Feature aggregation normally cost more memory and time during inference as features from multiple frames are needed. Combining the detection and the tracking is another direction worth attentions. Tracking could be an aid for the detection as it should propose better bounding boxes based on the detection results in the previous frames. As tracking looks for similarity between images, it is generally easier to track than to detect an object with deteriorated appearance between consecutive frames.
	
	The tracker design for the video object detection task is different from it is for the video object tracking task. There are several techniques in object tracking task have to be rejected. As the detection network requires powerful recognition ability, the features extracted are often heavier than those for tracking. It would be preferable if the tracker could re-use the features from the detection. However, such deep feature may not be compatible with some state of the art object trackers, e.g. siamese region proposal networks~\cite{siamrpn,siamrpn++} and fully-convolutional siamese networks~\cite{siamfc}. The performance would be heavily undermined by the different influence of paddings to the anchors in different spatial positions in a feature map. Another common practice in single-object tracking is to crop and resize the image patches of the template and the target to pre-defined sizes for valid network inputs. Such practice ensures that the features extracted are scale invariant to the real size of an object in the image. However, it is forbidden in the video object detection as it corrupts the features for object recognition.
	
	To improve the combination of the detection and the tracking, we propose a novel tracker for the video object detection task, which is compatible with the detection networks. The \textbf{main contributions} of this paper are following:
	\begin{itemize}
		\item We have created an object tracker for video object detection task. It is very light-weighted, memory efficient, computationally efficient and compatible to deep features of the existing modern detection networks.
		\item Our tracker can be inserted into a well trained image detector in a Plug \& Play style. Without harming the performance of the detector, the tracking functionality can be implanted into the model.
		\item Our new tracker performs in adaptive scales according to the size of the object being tracked, which makes it robust for tracking the object with large size variation in a video.
		\item We have designed a new video object detection pipeline to combine the advantages from both of the detection and the tracking. With better bounding box proposals and linkages through time, we improve the performance with better effectiveness and the efficiency.
	\end{itemize}
	
	\begin{figure*}
		\begin{center}
			\includegraphics[width=0.78\linewidth]{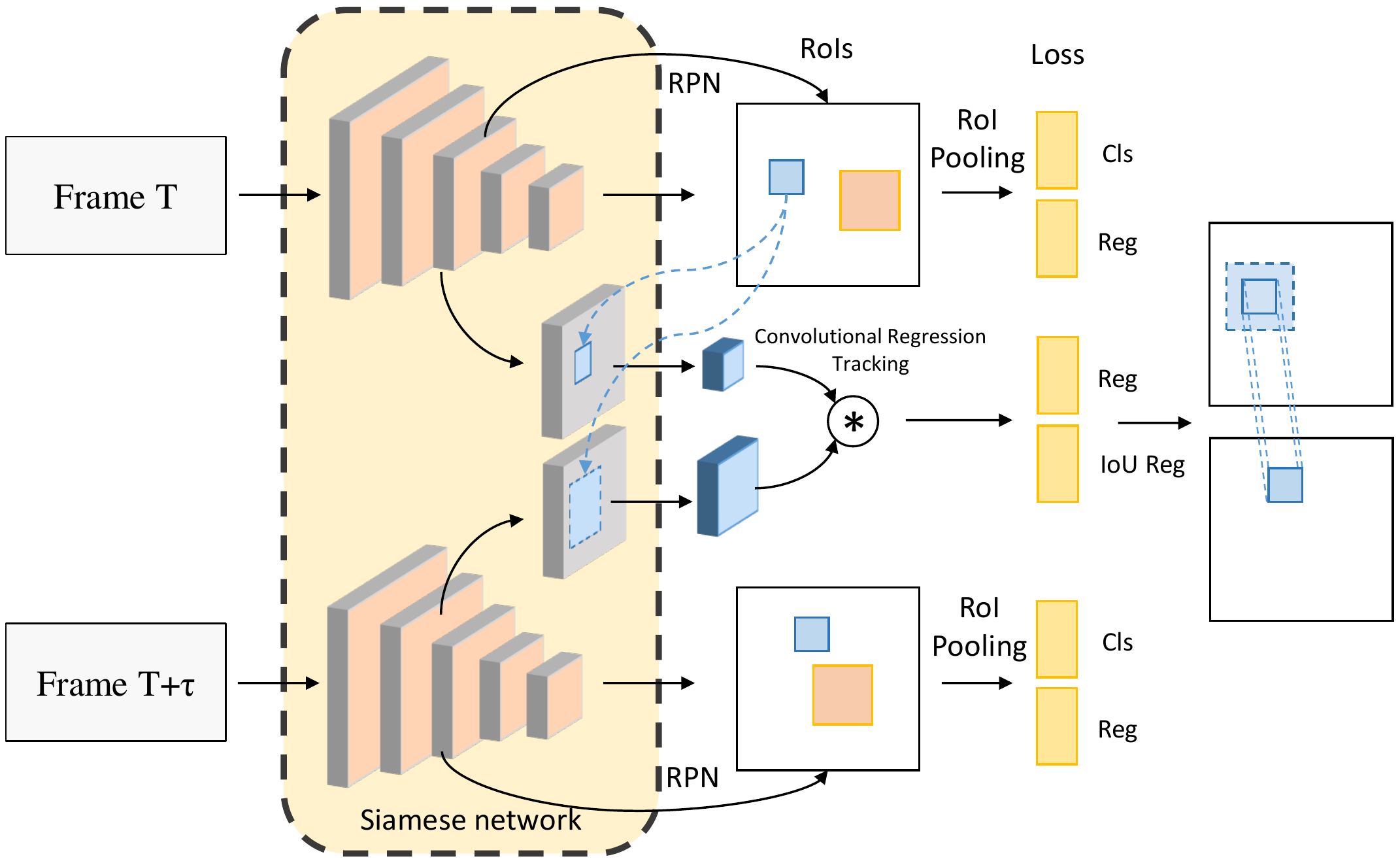}
		\end{center}
		\caption{Architecture of our video object detection network. Our Plug\&Play tracker reuses the features of the detection networks from both branches. The regional features within RoIs are pooled and sent to the tracker. The regional features from the two branches are convolved with each other for bounding box and IoU regression. (Details illustrated in Sec.~\ref{sec_overview}).}
		\label{fig:overview}
	\end{figure*}
	
	\section{Related work}
	Object detection in images is one of the fundamental tasks in computer vision. After the successful pioneer work~\cite{alexnet} with deep neural networks in object detection, a number of one-stage and two-stage object detector networks have been proposed~\cite{fast_rcnn,faster_rcnn,cascade_rcnn,panet,focal_loss,yolo,yolo9000,ssd,rfcn}.
	As an extension to object detection in images, object detection in videos has received much attention as well. Many methods have been proposed that introduce the idea of tracking, but very few achieve a real integration of detection and tracking networks. Instead, external trackers~\cite{TCNN} or optical flows~\cite{dff,FGFA,manet} or alternatives for tracking are required.
	
	Seq-NMS~\cite{seqnms} is a post-processing method for video object detection, which utilizes detection results from an object detector. Detections are linked through max score path finding under $0.5$ IoU score constraint between boxes in consecutive frames. Linked detections are re-scored afterwards. Such constraint is not optimal as it may not hold for objects in fast movement. 
	
	T-CNN~\cite{TCNN} proposes a deep learning framework that incorporates an external independent tracker~\cite{vt_fcn} to link the detections, which makes the pipeline slow. In~\cite{TPN} tubelet proposal network is utilized to propose tubelet boxes for multiple frames simultaneously. Boxes in the same tubelet are linked. ~\cite{track_cond} learns an additional feature embedding to help link the detected objects to the corresponding tracklets. ~\cite{scale_time} uses a scale-time lattice to accelerate the speed for video object detection. The temporal propagation for detection is inferred from the motion history images~\cite{MHI}. 
	
	D\&T~\cite{D&T} brings the tracking into the detection. By utilizing the feature map correlations between frames, the model learns a box regression model from one image to another. The tracked pairs are used to boost the linking score between image detections. However, D\&T~\cite{D&T} calculates the feature map correlations with a number of pre-defined position shifts and the feature map correlations have to be inferred for the whole image, which is very inefficient.
	
	Another direction for video object detection is through feature aggregation. By gathering clues from multiple consecutive frames, semantic information from both spatial and temporal domain can be extracted simultaneously to boost the detection performance.
	
	DFF~\cite{dff}, FGFA~\cite{FGFA}, MANet~\cite{manet} adopt optical flows to warp the features for alignment. Modern deep learning based optical flow models, such as FlowNet~\cite{flownet}, FlowNet2.0~\cite{flownet2} and LiteFlowNet~\cite{liteflownet} can process images in a very fast speed, which provide the chance for fast video object detector. However, optical flow models are normally trained with synthetic data and the performance of the tracking is limited by the domain discrepancy. STMN~\cite{stmn} aggregates the spatial-temporal memory for multiple frames according to the MatchTrans module, which is guided by the feature similarity between consecutive frames. STSN~\cite{stsn} directly extracts the spatially aligned features by using deformable convolutions~\cite{deformable_conv}. ~\cite{vod_PSLA} adopts progressive sparse local attentions to propagate the features across frames, while ~\cite{vod_ext_mem} utilizes explicit external memory to accumulate information over time. The temporal connection of these methods may not be accurate as there lacks an explicit learning process for tracking. ~\cite{vod_lrtr,vod_relation_distillation,vod_selsa} are more powerful feature aggregation methods as they distil semantic information from longer sequences.
	
	For tracking, siamese networks have received much attention recently. ~\cite{siamfc,cfnet} score the locations of objects by using feature correlation through a convolution operation between the template patch and the target patch. The idea is extended by~\cite{siamrpn,siamrpn++} with region proposal networks, which infer the object score and the box regression simultaneously for better box localization. GOTURN~\cite{goturn} adopts siamese network as feature extractor and uses fully connected layers to merge features for bounding box regression.
	
	Our work is inspired by the ideas in D\&T~\cite{D&T} and the siamese networks for tracking~\cite{goturn,siamfc,cfnet,siamrpn}.
	
	\section{Architecture Overview} \label{sec_overview}
	In this section, we will introduce an overview of our model structure. The goal of our model is to plug the tracking network into the detection network without harming the performance of the image detector. The architecture design is shown in Fig.~\ref{fig:overview}. Our model takes two consecutive frames with a gap of $\tau$ (1 for testing) as inputs $I^t,I^{t+\tau}\in\mathbb{R}^{H\times W\times3}$, followed by a siamese network for feature extraction. The two branches share the same weights to keep the identical feature extractors. To satisfy the need for object detection in complex scenes, we exploit powerful feature extraction backbones such as HRNet~\cite{hrnet} and ResNeXt~\cite{resnet,resnext}. The two models correspond to a light version detector and a heavy version detector. The extracted features from the siamese network are further sent to the detection branches and the tracking branch simultaneously. In a detection branch, regions of interest (RoIs) are proposed by the Region Proposal Network (RPN)~\cite{faster_rcnn}, followed by the RoI Pooling layer~\cite{fast_rcnn} to extract the features within each RoI. The RoI-wise features are exploited for object classification and bounding box regression, which is the same as the faster-rcnn~\cite{faster_rcnn}. One branch of the siamese network plus the detection branch forms a standard two-stage object detector. In the tracking branch, the novel scale-adaptive convolutional regression tracker is utilized to predict the bounding box transformation from the first frame to the second frame. The tracker utilizes regional features from the two branches of the siamese network based on the RoIs to be tracked. During training, the RoIs are generated from the ground truth bounding boxes, while in the testing phase, the RoIs are the detected objects. Besides the bounding box regression from the first image to the second image, the tracker has another IoU score regression branch to estimate the quality of the bounding box regression.
	
	\begin{figure}
		\begin{center}
			\includegraphics[width=0.95\linewidth]{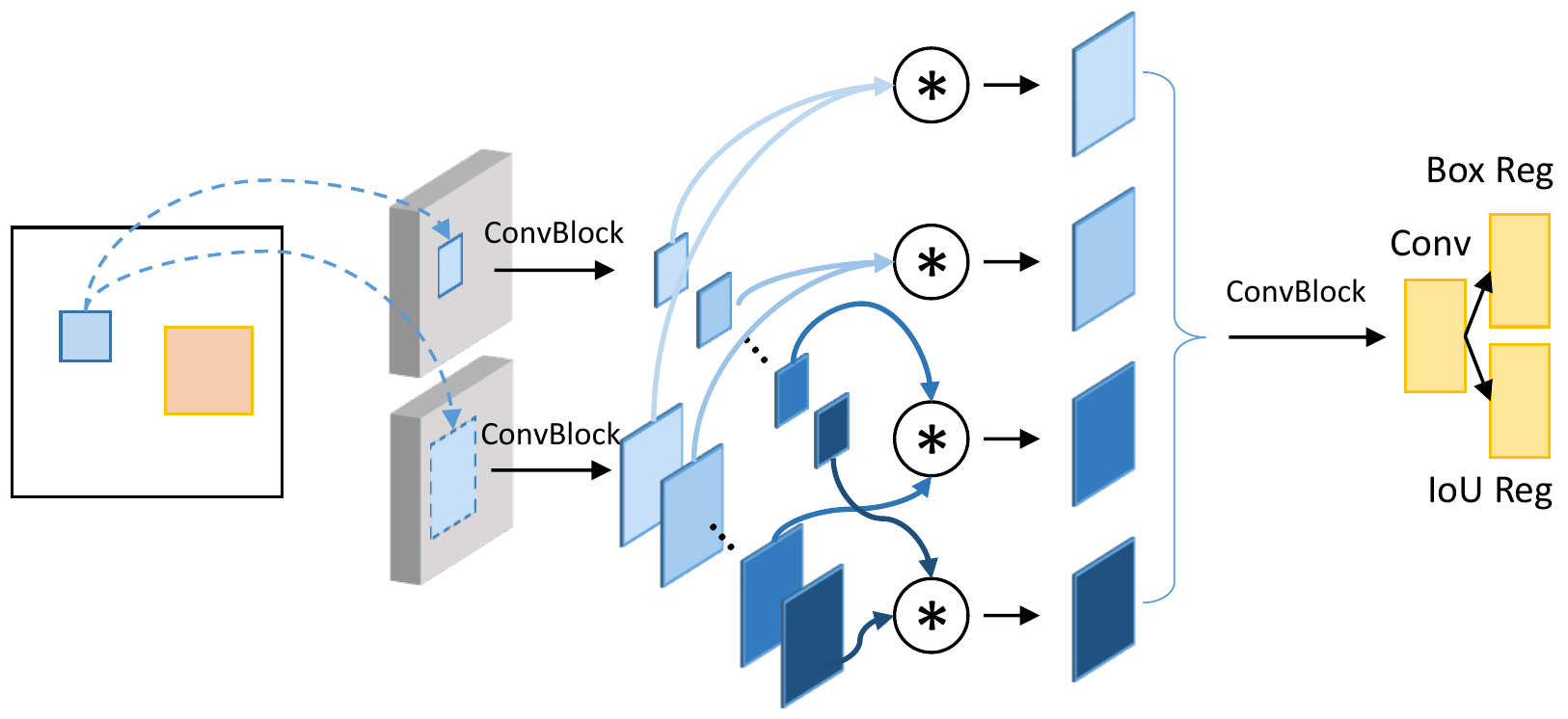}
		\end{center}
		\caption{Illustration of depth-wise feature correlation in our tracker. The bounding box of an object determines the location where the tracker acquires the local features. The features of each channel from the first branch is convolved with the features of the same channel from the second branch. Convolutional blocks are inserted to adjust the features, each of which is comprised of a convolutional layer, a batch normalization layer and a relu layer.}
		\label{fig:siam_conv}
	\end{figure}
	
	\section{Scale-adaptive convolutional regression tracker} \label{sec_traker}
	Many trackers for the video object tracking task crop and resize the image patches to fixed sizes according to the sizes of the objects to be tracked~\cite{siamrpn,siamrpn++,siamfc}. Such standard sizes for the network inputs make feature extraction invariant to the sizes of the objects. However, image patch cropping and resizing are not applicable to the detection network as there maybe multiple objects of different sizes. Instead of regularizing the size of the network inputs, we aim to extract scale-adaptive features for tracking by reusing features from the shared backbone, which augments the detector in a Plug \& Play style.
	
\begin{figure}
		\begin{center}
			\includegraphics[width=1.0\linewidth]{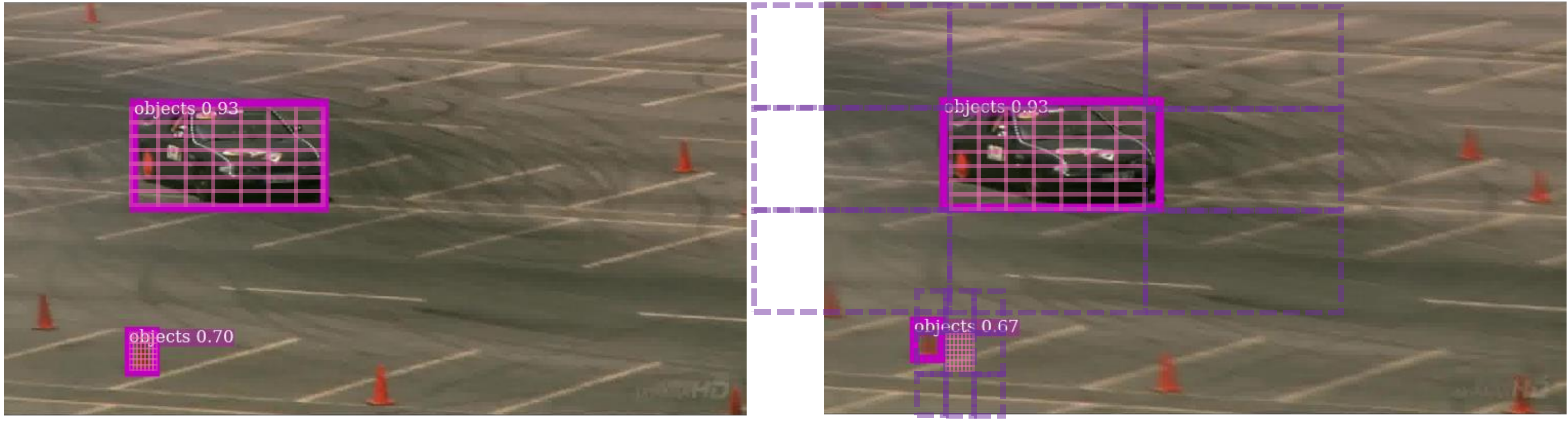}
		\end{center}
		\caption{Scale-adaptive tracking feature extraction. Here shows a real example of two objects being tracked. The $7\times7$ features are pooled within the object bounding boxes from the first image. The $21\times21$ features are pooled within the search area from the second image, which is $3$ times the size of the object.}
		\label{fig:track_illustrator}
	\end{figure}

	We extract regional features from both of the two branches based on the RoIs to be tracked. The RoI for the first branch marks the bounding box extent of an object. The width and height of the RoI bounding box are expanded $k$ times for the second branch with the center point and the aspect ratio fixed, which marks the local area of interest to search for the object in the second frame. $k$ is set to 3, indicating one object space to each side of the center object. RoIAlign~\cite{maskrcnn} is adopted to pool the features from the two branches of the siamese network. To keep the same scale of the pooled features, the pooled feature size from the second branch is also $k$ times the size of the first branch as shown in Fig.~\ref{fig:track_illustrator}. The features pooled from outside the range of the image are set to zero. We adopt the backbone features from multiple stages for RoIAlign pooling. Features from different stages are resized to the same intermediate size with max pooling and interpolation operations as in~\cite{librarcnn}. Instead of averaging, we concatenate the resized features. For HRNet-W32~\cite{hrnet} backbone, all stages are adopted. For ResNeXt101~\cite{resnext} backbone, features corresponding to middle $3$ stages are applied.
	
	D\&T~\cite{D&T} utilizes feature correlation as clues for tracking and encodes different translations into different channels. Our scale-adaptive tracker calculates feature correlation with a depth-wise convolution operation~\cite{siamrpn++} between two feature patches from the two branches of the siamese network. Fig.~\ref{fig:siam_conv} illustrates the process. Correlation of different translations are encoded into different spatial positions in our tracker. Convolutional blocks are inserted before and after the convolution for feature adjustment, each of which consists of a convolution layer, a batch normalization layer and a relu layer. The head of the tracker has a bounding box regression branch and an IoU score regression branch, which are two fully connected layers attached to a shared 2D convolution layer with $256$ filters. Sigmoid function is applied to normalize the IoU score prediction. As we adopt a class-agnostic regression tracking, the output dimensions are $4$ and $1$ respectively for each object. Smooth $L_1$ loss is utilized for regression~\cite{fast_rcnn}. Our tracker learns to predict the target bounding boxes directly so that it can rectify the boxes during tracking.
	
	\begin{figure}
		\begin{center}
			\includegraphics[width=1.0\linewidth]{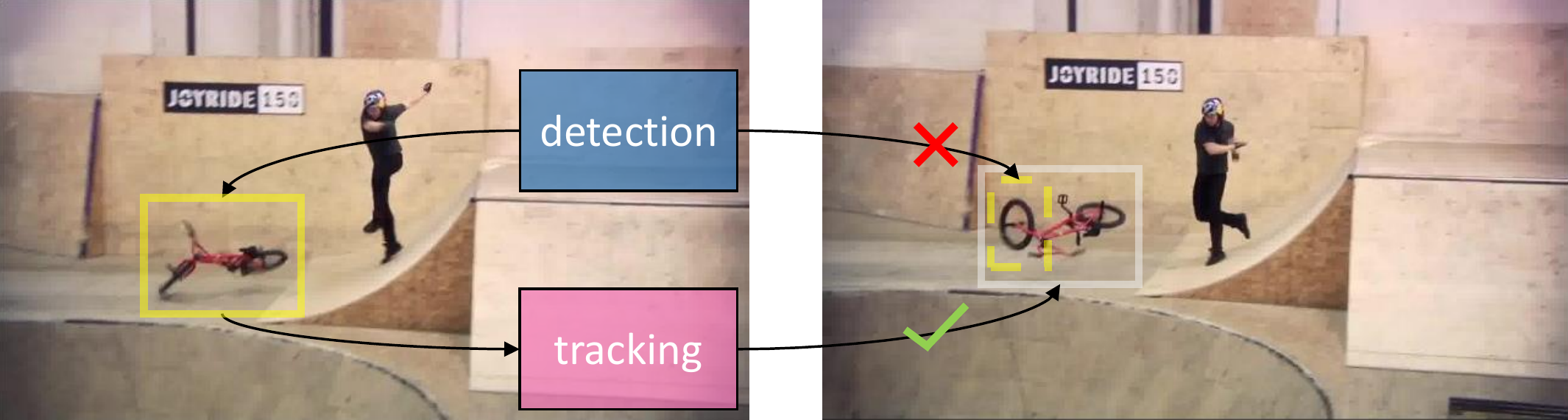}
		\end{center}
		\caption{Strategy for the bounding box selection. Bounding boxes can be inferred from both of the detection and the tracking process. We choose the tracked ones over the detected ones.}
		\label{fig:vod_nms}
	\end{figure}
	
	The learning targets are defined by the bounding boxes to be tracked $b^{t}=(b^{t}_x,b^{t}_y,b^{t}_w,b^{t}_h)$ in time $t$, the predicted bounding boxes $p^{t+\tau}=(p^{t+\tau}_x,p^{t+\tau}_y,p^{t+\tau}_w,p^{t+\tau}_h)$ in $t+\tau$ and the corresponding ground truth bounding boxes $g^{t+\tau}=(g^{t+\tau}_x,g^{t+\tau}_y,g^{t+\tau}_w,g^{t+\tau}_h)$ in $t+\tau$.
	The target for bounding box regression $\Delta^{t+\tau}=(\Delta^{t+\tau}_x,\Delta^{t+\tau}_y,\Delta^{t+\tau}_w,\Delta^{t+\tau}_h)$ is defined as,
	\begin{equation}\label{eq_target}
	\begin{split}
	\Delta^{t+\tau}_x&=\frac{g^{t+\tau}_x-b^{t}_x}{b^{t}_w},
	\Delta^{t+\tau}_y=\frac{g^{t+\tau}_y-b^{t}_y}{b^{t}_h}\\
	\Delta^{t+\tau}_w&=\ln\frac{g^{t+\tau}_w}{b^{t}_w},
	\Delta^{t+\tau}_h=\ln\frac{g^{t+\tau}_h}{b^{t}_h}\\
	\end{split}
	\end{equation}
	The target for IoU score regression is calculated by $b^{t+\tau}_{score}=IoU(p^{t+\tau}, g^{t+\tau}),b^{t+\tau}_{score}\in[0,1]$. The predicted bounding boxes are inferred through predicted bounding box regression $\widehat{\Delta}^{t+\tau}=(\widehat{\Delta}^{t+\tau}_x,\widehat{\Delta}^{t+\tau}_y,\widehat{\Delta}^{t+\tau}_w,\widehat{\Delta}^{t+\tau}_h)$. $p^{t+\tau}$ can be calculated as,
	\begin{equation}\label{eq_pred}
	\begin{split}
	p^{t+\tau}_x&=\widehat{\Delta}^{t+\tau}_x b^{t}_w+b^{t}_x,
	p^{t+\tau}_y=\widehat{\Delta}^{t+\tau}_y b^{t}_h+b^{t}_y\\
	p^{t+\tau}_w&=\exp(\widehat{\Delta}^{t+\tau}_w)b^{t}_w,
	p^{t+\tau}_h=\exp(\widehat{\Delta}^{t+\tau}_h)b^{t}_h\\
	\end{split}
	\end{equation}
	
	Our feature extraction design for tracking has several advantages. First, only resized local features are utilized, which makes the tracker memory efficient. Second, features extracted from the two branches are of the same scale and are in scale to the size of an object. The feature correlation can be calculated in an adaptive scale and range, which aligns with the scale-adaptive learning target for bounding box regression. Finally, the tracker is light-weighted as it reuses the features from the backbone network.
	
	\section{Combination of detection and tracking}
	Our model has the functionality for both of the detection and the tracking. In this section, we introduce how the detection and the tracking are combined to solve the video object detection task. The aim of detection is to find the newly appeared objects in an image, while the tracking is to better localize the objects across the frames. 
	
	\textbf{Detect to track.} 
	The two-stage detection network has two object proposal stages, the RoI proposals by RPN and the object detections by R-CNN. We base our tracking on top of the object detections instead of the RoI proposals as the R-CNN provides much cleaner and more accurate objects for tracking. The design of RPN is to provide proposals with high enough recall rate and it is not time or memory efficient to track hundreds of RoIs. Setting a score threshold for foreground proposal selection would not be good either as the scores and bounding boxes would not be accurate enough due to the anchor based design. The detection results from R-CNN is more robust. If an object is identified in an image with a high enough class score ($0.03$ in our case), it would be a candidate for tracking in the next image.
	
	\textbf{Track to detect.} 
	Tracking could aid the detection by providing better object boxes. We first filter the tracked objects. If the predicted tracking IoU score is too small for an object ($0.5$ in our case), the object is discarded. As multiple objects may overlap with each other during tracking, we apply a non-maximum suppression ($0.7$ IoU threshold in our case) to the tracked boxes based on the IoU score. The selected objected are used in the video object detection.
	A tracking first detection (TFD) strategy is applied. When an object is identified by both of the detection and the tracking, we choose the tracked one over the detected one as shown in Fig.~\ref{fig:vod_nms}. The non-maximum suppression in detection favours the objects with higher class scores instead of the objects with more accurate bounding boxes. We favour the better bounding boxes by adopting the TFD strategy to help acquire better object linking across frames. Only the newly detected objects that have IoUs with the existing tracked ones lower than a threshold $T^{nms}_{merge}$ are reserved. During the inference across all the image frames in a video, the tracked boxes are saved. Detections are further refined by averaging re-scoring and non-maximum suppression as in~\cite{seqnms}. 
	
	\section{Experiments}
	\subsection{Dataset and evaluation}
	Our method is evaluated on the ImageNet~\cite{imagenet} object detection from video (VID) dataset. There are 3862 training and 555 validation videos with objects from 30 classes labelled for the task. The ground truth annotations contain the bounding box, the class ID and the track ID for each object. The performance of the algorithm is measured with mean average precision (mAP) score over the 30 classes on the validation set as it is in~\cite{FGFA,manet,D&T,TCNN,TPN,stmn,stsn}. In addition to the ImageNet VID datset, the ImageNet object detection (DET) dataset has 200 classes, which include all the 30 classes in the ImageNet VID dataset as well. We follow the common practise by utilizing the data from both of the DET and the VID dataset~\cite{FGFA,manet,D&T,TCNN,TPN,stmn,stsn}.
	
	\subsection{Configurations}
	\textbf{Image training.} In the first stage, we train the detection parts of our video object detector in the same way as a standard object detector. The training samples are from both of the DET and the VID dataset. To balance the classes in the DET dataset, we sample at most $2K$ images from each of the $30$ categories to get our DET image set ($53K$ images). To balance the VID videos, which have large sequence length variations, we evenly sample $15$ frames from each of the video sequence to get our VID image set ($57K$ images). The combined DET+VID image set is used for detector training. 
	We apply SGD optimizer with a learning rate of $10^{-3}$ for the first $90K$ iterations and $10^{-4}$ for the last $45K$ iterations. The training batch is set to $8$ images that are distributed among $4$ gpus. In both of the training and the testing, we apply a single scale with the shorter dimension of the images to be 600 pixels. During training, only random left-to-right flip is used for data augmentation~\footnote{Please check the supplementary material for detailed configurations.}.
	
	\textbf{Video training.}
	In this stage, we further train our tracking parts with the image pairs from the ImageNet VID dataset. We randomly select two consecutive images with a random temporal gap from $1$ to $9$ frames. As there is no causal reasoning involved, we randomly reverse the sequence order to gain more variety of translations. The RoIs for tracking $R=(R_x,R_y,R_w,R_h)$ are generated by resizing and shifting the ground truth bounding boxes $g=(g_x,g_y,g_w,g_h)$ as in Eq.~\ref{eq_track_rois}. The coefficients $\delta=(\delta_x,\delta_y,\delta_w,\delta_h)$ are sampled from uniform distributions $U$. $\delta_x,\delta_y\in U[-1.0,1.0]$ and $\delta_w,\delta_h\in U[0.5,1.5]$. For each ground truth object, we sample $256$ RoIs and randomly select $128$ RoIs from those satisfying the constraint of $IoU(R,g)<0.5$.
	\begin{equation}\label{eq_track_rois}
	\begin{aligned}
	R_x&=\delta_x g_w+g_x&, R_y&=\delta_y g_h+g_y,\\
	R_w&=\delta_w g_w&, R_h&=\delta_h g_h,
	\end{aligned}
	\end{equation}
	We freeze the backbone parts to train the tracking parts only in order to retain the accuracy for detection. RPN and R-CNN are not included either. The model is trained with SGD optimizer with a learning rate of $10^{-3}$ for the first $80K$ iterations and $10^{-4}$ for the next $40K$ iterations. We apply a batch of $16$ image pairs for training that are distributed among $4$ gpus. The images are resized to the same single scale as the image training step.
	
	\textbf{Testing.}
	In the testing stage, we select the detected objects with the class scores higher than $0.03$ and apply an IoU threshold of $0.45$ for the final detection output. The single scale testing is utilized with the shorter side of images resized to 600 pixels.
	
	\textbf{Implementations.}
	Our model is implemented with pytorch~\cite{pytorch} and integrated with MMDetection~\cite{mmdetection}.
	
	\subsection{Results}
	We compare several major competitive video object detection algorithms in Tab.~\ref{tab_cmp}. FGFA~\cite{FGFA} and MANet~\cite{manet} utilize optical flow to guide linking between frames, but they cannot achieve a good balance between the mAP score and the speed. STMN~\cite{stmn} and STSN~\cite{stsn} aggregate information from multiple frames, which limits the speed greatly. By bringing a very light-weighted tracker into the model, our light and heavy models achieve scores comparable to some of the state-of-the-art methods.
	\begin{table}
		\small
		\begin{center}
			\begin{adjustbox}{width=1.0\linewidth}
				\begin{tabular}{l|c|c|c}
					\hline
					Methods              & Temporal link    & mAP (\%)   & FPS\\
					\hline
					FGFA~\cite{FGFA}     & optical flow     & 78.4	   & 1.15\\
					FGFA+~\cite{FGFA}    & optical flow     & 80.1	   & 1.05\\
					MANet~\cite{manet}   & optical flow     & 78.1	   & 4.96\\
					MANet+~\cite{manet}  & optical flow     & 80.3	   & -\\
					STMN~\cite{stmn}     & STMM             & 80.5	   & 1.2\\
					STSN~\cite{stsn}     & DCN              & 78.9     & -\\
					STSN+~\cite{stsn}    & DCN              & 80.4     & -\\
					D\&T~\cite{D&T}      & box regression   & 79.8	   & 7.09\\
					PSLA+~\cite{vod_PSLA}& attention        & 81.4     & 5.13\\
					EMN~\cite{vod_ext_mem}& memory          & 79.3     & 8.9\\
					EMN+~\cite{vod_ext_mem}& memory         & 81.6     & -\\
					Ours(HRNet-w32)	     & box regression   & 78.6	   & 11\\
					Ours(ResNeXt101*)    & box regression   & 81.1     & 5.6\\
					\hline
				\end{tabular}
			\end{adjustbox}
		\end{center}
		\caption{Comparisons among different video object detection methods. + stands for Seq-NMS~\cite{seqnms}. - stands for not provided. * means with FPN~\cite{FPN} and DCN~\cite{deformable_conv}.} 
		\label{tab_cmp}
	\end{table}

	\subsection{Ablation study}
	To test the effectiveness of bringing the tracking into the model, we perform ablation study by gradually adding the components into the model. We start with the per-image detection model. The faster-RCNN with HRNet backbone is adopted as the basic model. We first test the standard detection result without any aid from the tracking. The performance score is shown in Tab.~\ref{tab_ablation}. The per-image detection achieves a decent mAP score of $73.2\%$. We further add Seq-NMS~\cite{seqnms} to see the performance of linking and re-scoring based solely on the detection results. As in~\cite{seqnms}, we set the IoU threshold for boxes linking constraint to be $0.5$ and IoU threshold for detection NMS to be $0.45$. The average precision scores improve for all the categories and the mAP score has increased by $2.3\%$. We further add tracking into our model by adopting our TFD video object detection strategy. During the training of the tracking modules, we freeze the parameters of the feature extraction backbone, the RPN and the R-CNN in order to control the performance of the object detector. We compare two values for merging NMS IoU threshold $T^{nms}_{merge}$, $0.3$ and $0.7$ (marked as TFD(0.3) and TFD(0.7) in Tab.~\ref{tab_ablation}). The mAP scores have increased another $2.3\%$ and $3.1\%$ respectively. The mAP score is higher with $T^{nms}_{merge}=0.7$, showing that the video object detector still benefits more from the denser object proposals. The TFD strategy is very effective to help improve the quality of linking and re-scoring. By now, we have improved the performance of the object detector by a large margin (+$5.4\%$ mAP) without modifying any parameter of an object detector. With heavier ResNeXt101~\cite{resnext} backbone plus FPN~\cite{FPN} and DCN~\cite{deformable_conv}, the TFD+Seq-NMS improves the detector by a large margin (+$4.2\%$ mAP) from $76.2\%$ to $80.4\%$.
	Seq-NMS links boxes across frames under constraint $IoU(b^t, b^{t+1})>0.5$, which fails if there is large position translation between consecutive frames. We improve the constraint to be $IoU(p^{t+1}, b^{t+1})>0.5$, where $p^{t+1}$ is the predicted box from $b^{t}$ by the tracker. The improved re-scoring method (Seq-Track-NMS) provides another $0.7\%$ mAP score boost.
	
	\begin{table}
	\small
		\begin{center}
			\begin{adjustbox}{width=1.0\linewidth}
				\begin{tabular}{l|c}
					\hline
					Methods &mAP (\%)\\
					\hline
					HRNet-w32                            &73.2\\
					HRNet-w32+Seq-NMS                    &75.5\\
					HRNet-w32+TFD(0.3)+Seq-NMS           &77.8\\
					HRNet-w32+TFD(0.7)+Seq-NMS           &78.6\\
					ResNeXt101*                          &76.2\\
					ResNeXt101*+TFD(0.7)+Seq-NMS         &80.4\\
					ResNeXt101*+TFD(0.7)+Seq-Track-NMS   &81.1\\
					\hline
				\end{tabular}
			\end{adjustbox}
		\end{center}
		\caption{Ablation study of our method. TFD stands for tracking first detection. * stands for with FPN~\cite{FPN} and DCN~\cite{deformable_conv}.}
		\label{tab_ablation}
	\end{table}
	
	\begin{figure}
		\begin{center}
			\includegraphics[width=1.0\linewidth]{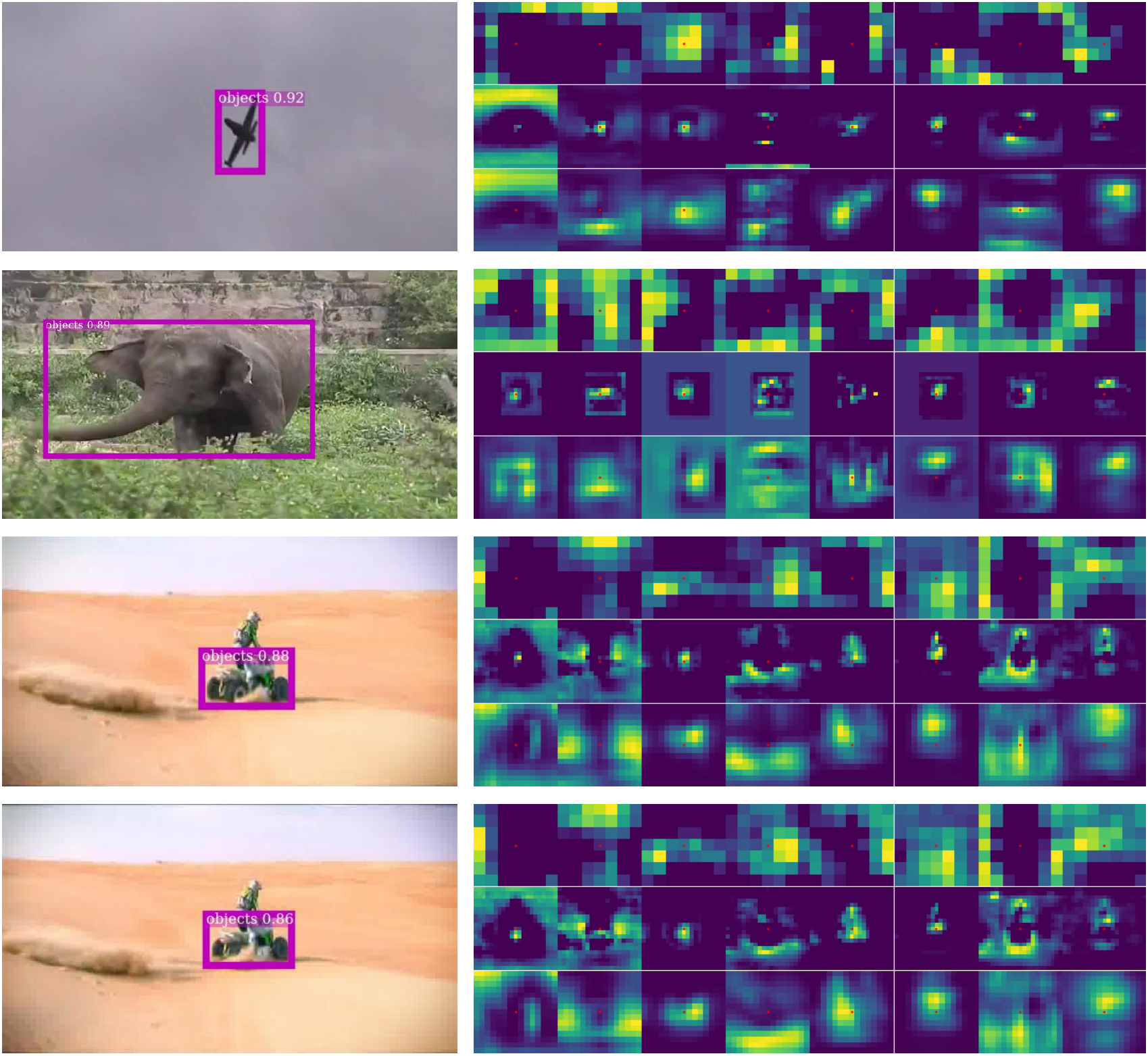}
		\end{center}
		\caption{Correlation feature visualization. Images on the left show the objects being tracked. Feature maps on the right are the correlation features and the output features. For each image, the upper two rows are the correlation features from the two branches while the bottom row shows the convolution output. The randomly selected 8 channels are ordered in 8 columns. The center of a feature map is marked with a red dot for spatial reference.}
		\label{fig:feat_vis}
	\end{figure}
	\subsection{Visualization of the correlation features}
	To make sure that the tracker utilizes the features from both of the branches rather than being dominated by one branch, we randomly sample the same 8 channels from the features before and after the correlation operation for visualization to check how the features for tracking are produced. The features are shown in Fig.~\ref{fig:feat_vis}. The examples are air-plane, elephant and motorcycles. The two motorcycle examples are two consecutive images, where the second image shows an inverse sequence order for tracking. It should be noted that the template and the target patches are encoded in the same scale but of different sizes. It is interesting to notice that the center of mass of the template features are shifted to all different positions. The features of different objects are distinctive and the correlation output is greatly affected from both of the branches. By examining the examples of the two motorcycles, which have opposite moving patterns, it could be seen that the features from the target branch determine the tracking prediction while the features from the template branch are more similar. In conclusion, the tracking prediction is mainly affected by the relative position for an object. The template features of different objects would encode the target features differently. The features from both of the branches affect the tracking prediction in the same time.
	
	\subsection{The efficiency of time and memory}
	In this section, the efficiency of time and memory are examined. We take HRNet-w32 as an example for the following analysis. All the experiments are conducted with a single Titan X (Pascal) GPU during testing stage. The design of our tracking module is very light-weighted in both of the time and the memory. We first test the time efficiency of different components in our model. The approximate time costs are reported in Tab.~\ref{tab_time}. Our additional tracker is very efficient and costs only an extra time of $3$ ms, which is lighter than the RPN or the R-CNN.
	
	We further examine the running GPU memory cost of our model. We compare the GPU memory consumption with and without the tracker as shown in Tab.~\ref{tab_memory}. Our tracker requires only another 60 MB to run, which is very memory efficient.
	
	\begin{table}
		\begin{center}
			\begin{adjustbox}{width=1.0\linewidth}
				\begin{tabular}{c|c|c|c|c}
					\hline
					Components&Backbone&RPN&R-CNN&Tracker\\
					\hline
					Time (ms) &43 &10 &7 &3\\
					\hline
				\end{tabular}
			\end{adjustbox}
		\end{center}
		\caption{Time cost comparison of different components.}
		\label{tab_time}
	\end{table}

	\begin{table}
		\begin{center}
				\begin{tabular}{c|c|c}
					\hline
					Model      & Detector& Detector+Tracker\\
					\hline
					Memory (GB)& 1.59    & 1.65\\
					\hline
				\end{tabular}
		\end{center}
		\caption{Memory cost comparison of different models.}
		\label{tab_memory}
	\end{table}

	\begin{table}
		\small
		\begin{center}
			\begin{adjustbox}{width=1.0\linewidth}
				\begin{tabular}{c|c|c|c|c}
					\hline
					Backbone& Detector& TFD  & Seq-NMS & FPS\\
					\hline
					HRNet-w32&{\cmark}&{\xmark}  &{\xmark}  &15\\
					
					HRNet-w32&{\cmark}& {\cmark} & {\xmark} &12\\
					
					HRNet-w32&{\cmark}& {\cmark} & {\cmark} &11\\
					\hline
					ResNeXt101*&{\cmark}& {\xmark} & {\xmark} &6.9\\
					
					ResNeXt101*&{\cmark}& {\cmark} & {\cmark} &5.6\\
					\hline
				\end{tabular}
			\end{adjustbox}
		\end{center}
		\caption{Time efficiency comparison of the pipelines. * stands for with FPN~\cite{FPN} and DCN~\cite{deformable_conv}.}
		\label{tab_model_time}
	\end{table}

	For the video object detection, we examine the running speed of our pipeline and analyse the effect of different components. The speed is measured in frames per second (FPS). The detector with HRNet-w32 backbone runs at $15$ FPS and our tracker embedded pipeline runs at $12$ FPS. The additional Seq-NMS slows down our pipeline slightly to $11$ FPS. As our detector with ResNeXt101 backbone is combined with FPN~\cite{FPN} and DCN~\cite{deformable_conv}, it runs relatively slower but still faster than methods like FGFA+~\cite{FGFA}, STMN~\cite{stmn} and MANet+~\cite{manet}.
	
	\begin{figure*}
		\begin{center}
			\includegraphics[width=0.8\linewidth]{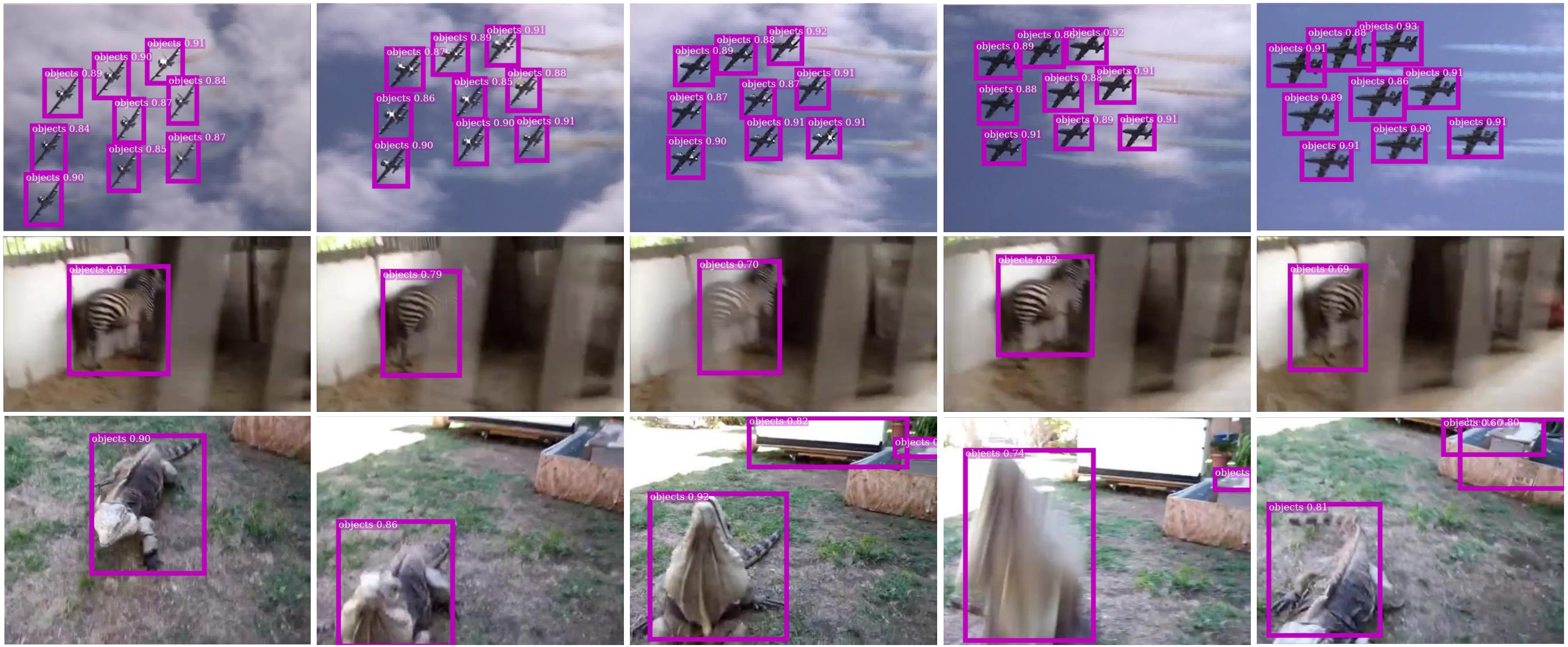}
		\end{center}
		\caption{Tracking examples by our Plug \& Play tracker. Our tracker can track single or multiple objects and can handle problems like motion blur, partial occlusion and rare object poses. The objects are labelled with IoU regression scores.}
		\label{fig:track_egs}
	\end{figure*}
	\begin{figure*}
		\begin{center}
			\includegraphics[width=0.8\linewidth]{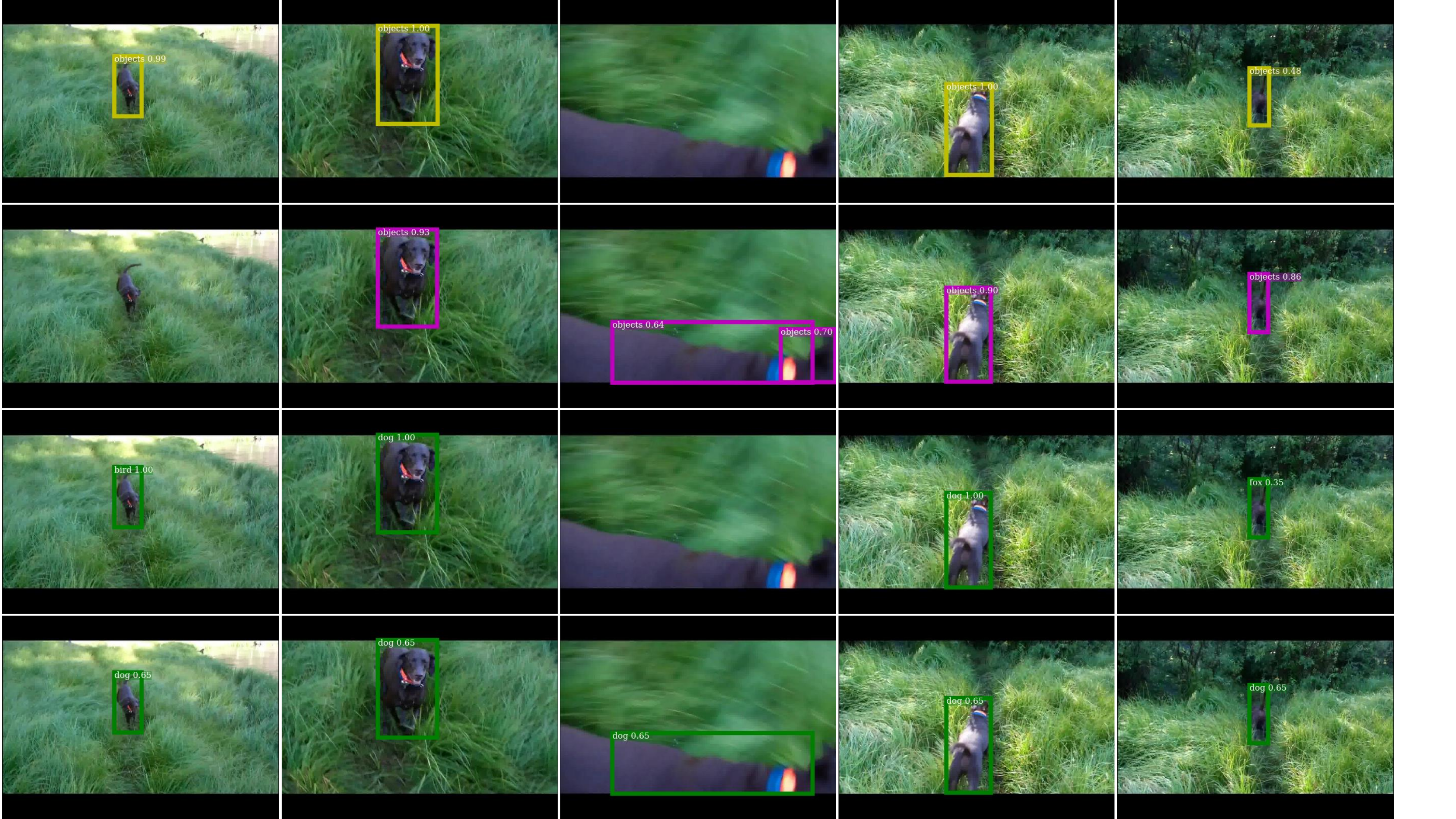}
		\end{center}
		\caption{A video object detection example. The $0_{th}, 20_{th}, 40_{th}, 60_{th}$, and $80_{th}$ images are shown. The first row shows the proposed object from the detector. The second row shows the proposed object from the tracker. The third row shows the detection from the combined proposals. The fourth row shows the re-scored detection results. Only the objects with scores above $0.2$ are shown.}
		\label{fig:det_egs}
	\end{figure*}
	
	\subsection{Qualitative results}
	Our tracker learns to track objects in different circumstances. The tracking examples of motion blur, rare pose, partial occlusion and multiple objects are shown in Fig.~\ref{fig:track_egs}. By incorporating tracking into the detection, our video object detector can achieve very long term detection consistency across frames. Fig.~\ref{fig:det_egs} shows an example of a running dog in a long sequence. This example clearly shows how the detector and the tracker collaborate. The detector finds new objects while the tracker follows the objects. The tracker provides better boxes for linking and re-scoring. Without re-scoring, the detection could be wrong or weak as shown in the third row (The dog is wrongly classified as a bird in the first column or a fox in the last column). After re-scoring, the long term detection consistency can be achieved as shown in the fourth row~\footnote{More qualitative results are shown in the supplementary material.}.
	
	\section{Conclusion}
	We have proposed a Plug \& Play convolutional regression tracker that augments the image detectors for the object detection in videos. The tracker makes use of the deep features from the image detectors for tracking with very little extra memory and time cost. The light-weighted tracker can track a single object or multiple objects, and handle problems e.g. motion blur, defocus, rare poses or partial occlusions. With our tracking first detection strategy and the improved Seq-NMS~\cite{seqnms} linking and re-scoring method, the performance of our detector improves by a large margin. 
	
	{\small
		\bibliographystyle{ieee_fullname}
		\bibliography{egbib}
	}
	
	\section*{Appendix}
	\subsection*{Additional configuration details.}
	\noindent
	\textbf{Features extraction for tracking}:
	The backbone features from the light HRNet-w32~\cite{hrnet} model and the FPN~\cite{FPN} features from the heavy ResNeXt101~\cite{resnext} model are utilized as input for tracker. Features are resized and concatenated. Our ResNeXt101 model has the cardinality of $32$ and the base width of $4$.
	\begin{figure}[h]
		\centering
		\includegraphics[width=0.9\linewidth]{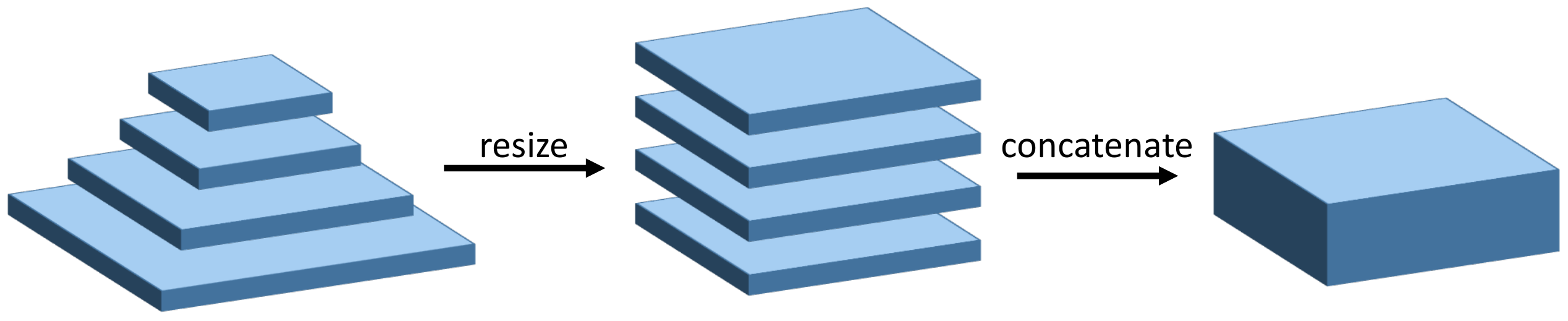}
		\caption{The feature from the HRNet-w32 backbone for tracking. The features are spatially resized to the size of the features in the second stage.}
		\includegraphics[width=0.9\linewidth]{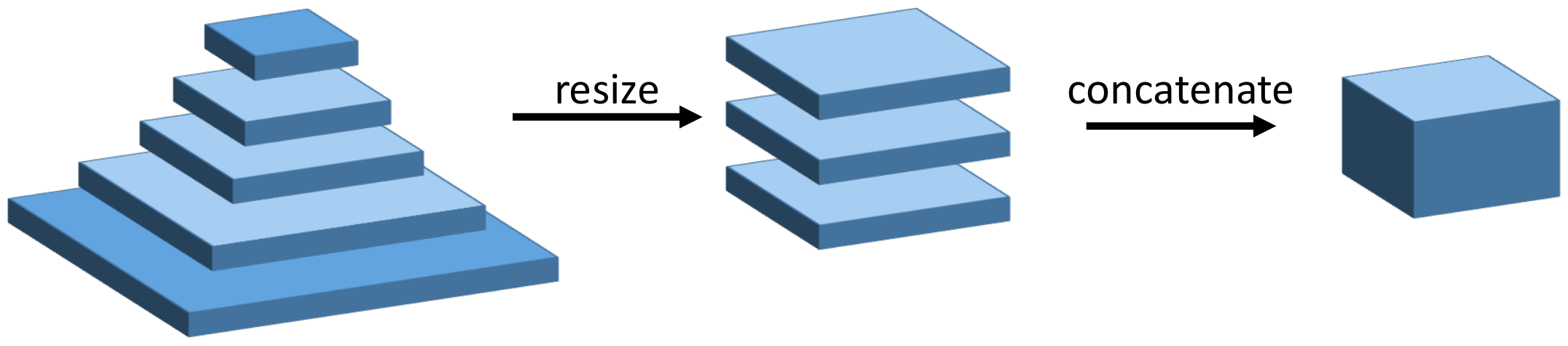}
		\caption{The features from the FPN (coupled with ResNeXt101) for tracking. The features from the middle $3$ stages are resized to the size of the features in the third stage.}
	\end{figure}

	\noindent
	\textbf{Region Proposal Network (RPN)}: 
	The anchors in RPN have $3$ aspect ratios ($0.5,1,2$) spanning $5$ scales ($32,64,128,256,512$). For the ResNeXt model with FPN, $5$ scales are distributed to the $5$ stages in the FPN pyramid.
	
	\noindent
	\textbf{RoIAlign Pooling}: The RoI pooling module has a size of $7\times7$. For the tracker, the pooled feature is the average of the features in a bin. For the detector, the pooled feature is the average of the nearest $4$ features.
	
	\noindent
	\textbf{R-CNN}: 
	The R-CNN has a bounding box regression branch and a logistic regression branch for classification. The two branches have $2$ shared fully connected layers with $1024$ filters, attached with $1$ fully connected layer in each branch for their own purpose.
	
	\noindent
	\textbf{Deformable-ConvNets (DCN)}: 
	The DCN~\cite{deformable_conv} with $32$ groups and modulation~\cite{dcn2} is applied for stage $3$,$4$,$5$ in ResNeXt~\cite{resnext} model.
	
	\noindent
	\textbf{Image training}: 
	Online hard example mining (OHEM)~\cite{OHEM} is utilized for R-CNN training.
	

	\subsection*{Detailed results}
	We additionally provide detailed mAP scores for ablation study as shown in Table~\ref{tab_ablation}.
	
	\subsection*{More qualitative examples}
\begin{figure*}
		\centering
		\includegraphics[width=\linewidth]{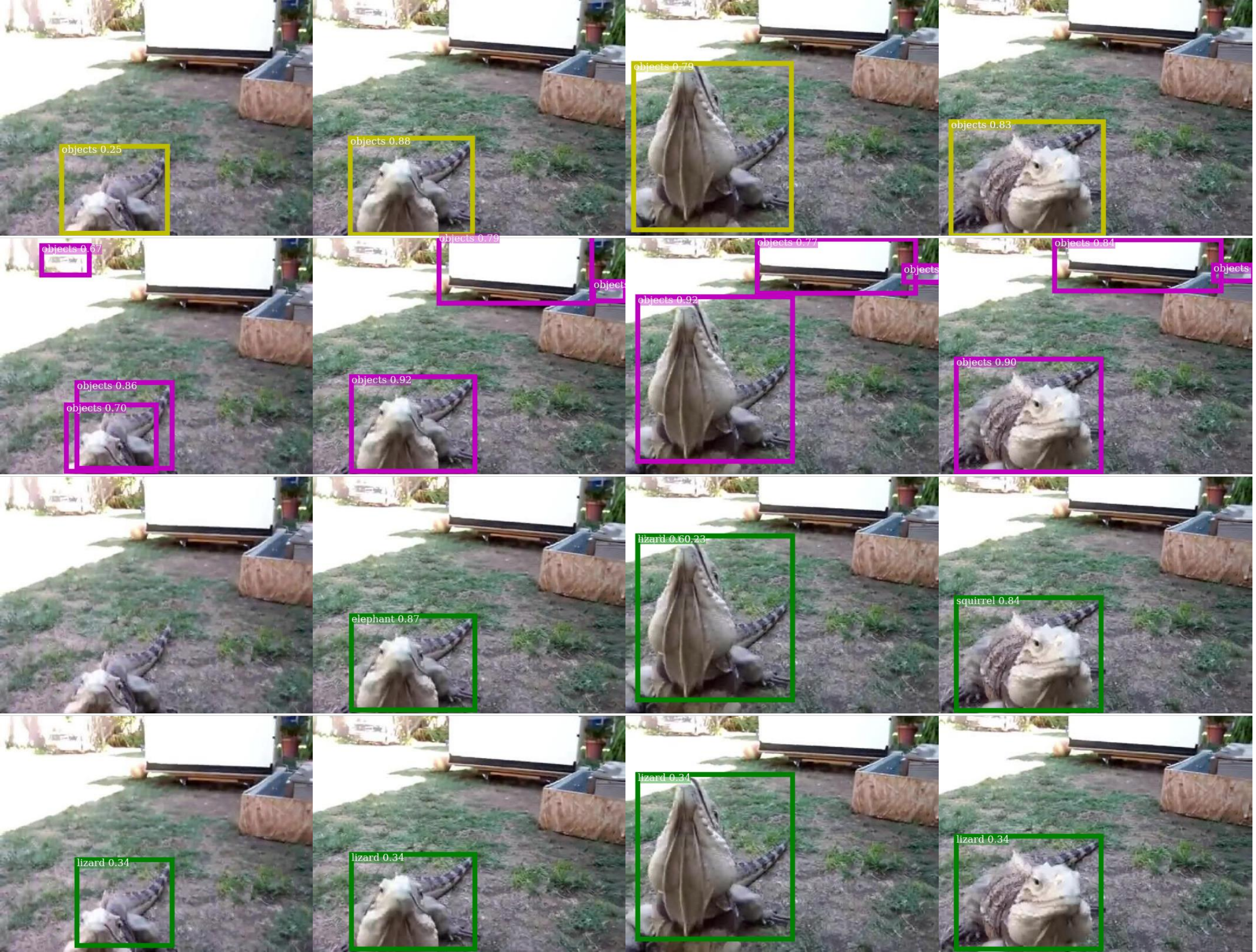}
		\caption{Example of lizard. The first row shows the proposed
			object from the detector. The second row shows the proposed object from the tracker. The third row shows the detection from the combined proposals. The fourth row shows the re-scored detection results. Only the objects with scores above 0.2 are shown. The weak or wrong detections shown in the third row are rectified as shown in the fourth row.}
		\label{fig_1}
	\end{figure*}

\begin{figure*}
		\centering
		\includegraphics[width=\linewidth]{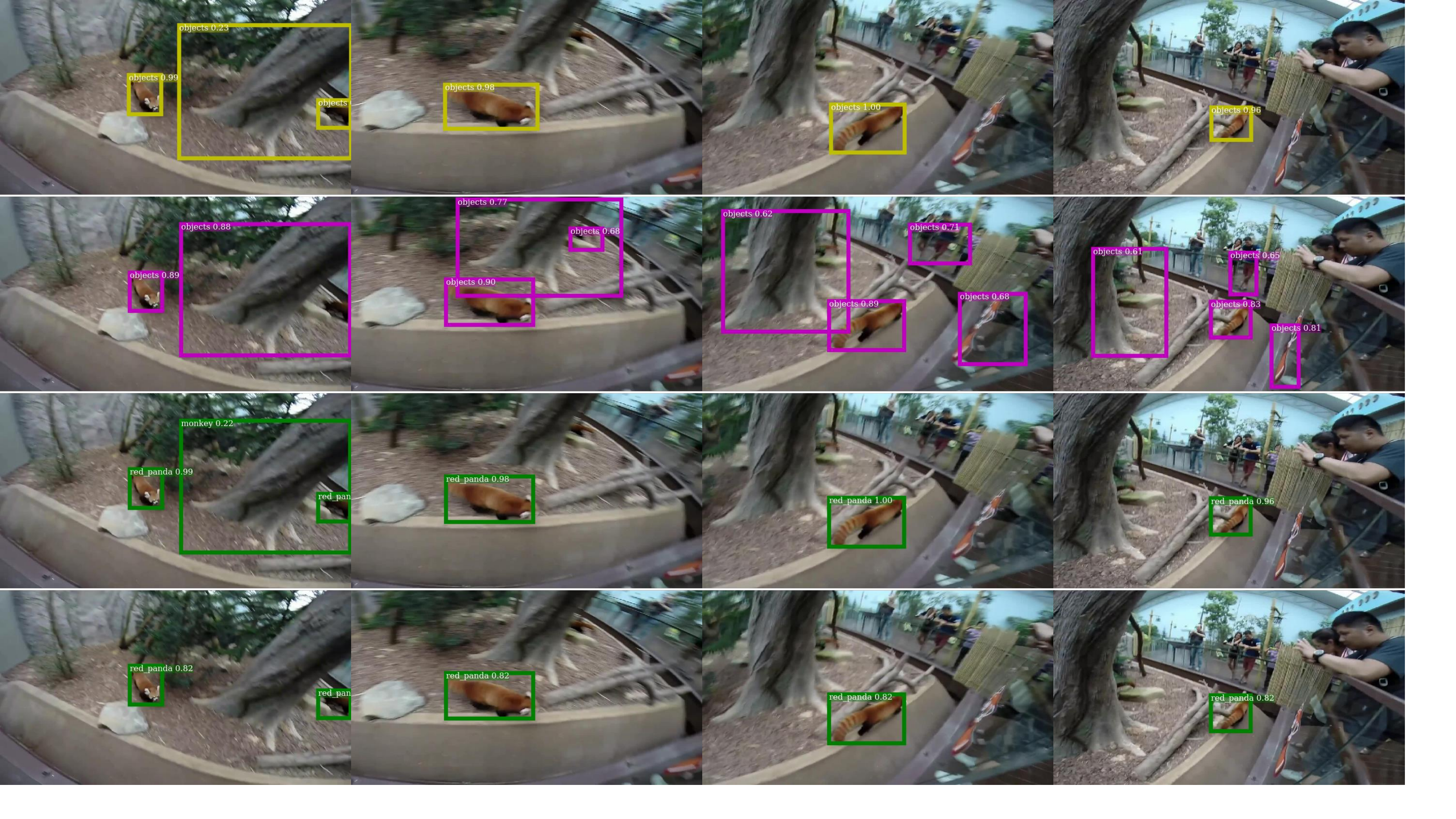}
		\caption{Example of red panda. The first row shows the proposed
			object from the detector. The second row shows the proposed object from the tracker. The third row shows the detection from the combined proposals. The fourth row shows the re-scored detection results. Only the objects with scores above 0.2 are shown.}
		\label{fig_2}
	\end{figure*}

\begin{figure*}
		\centering
		\includegraphics[width=\linewidth]{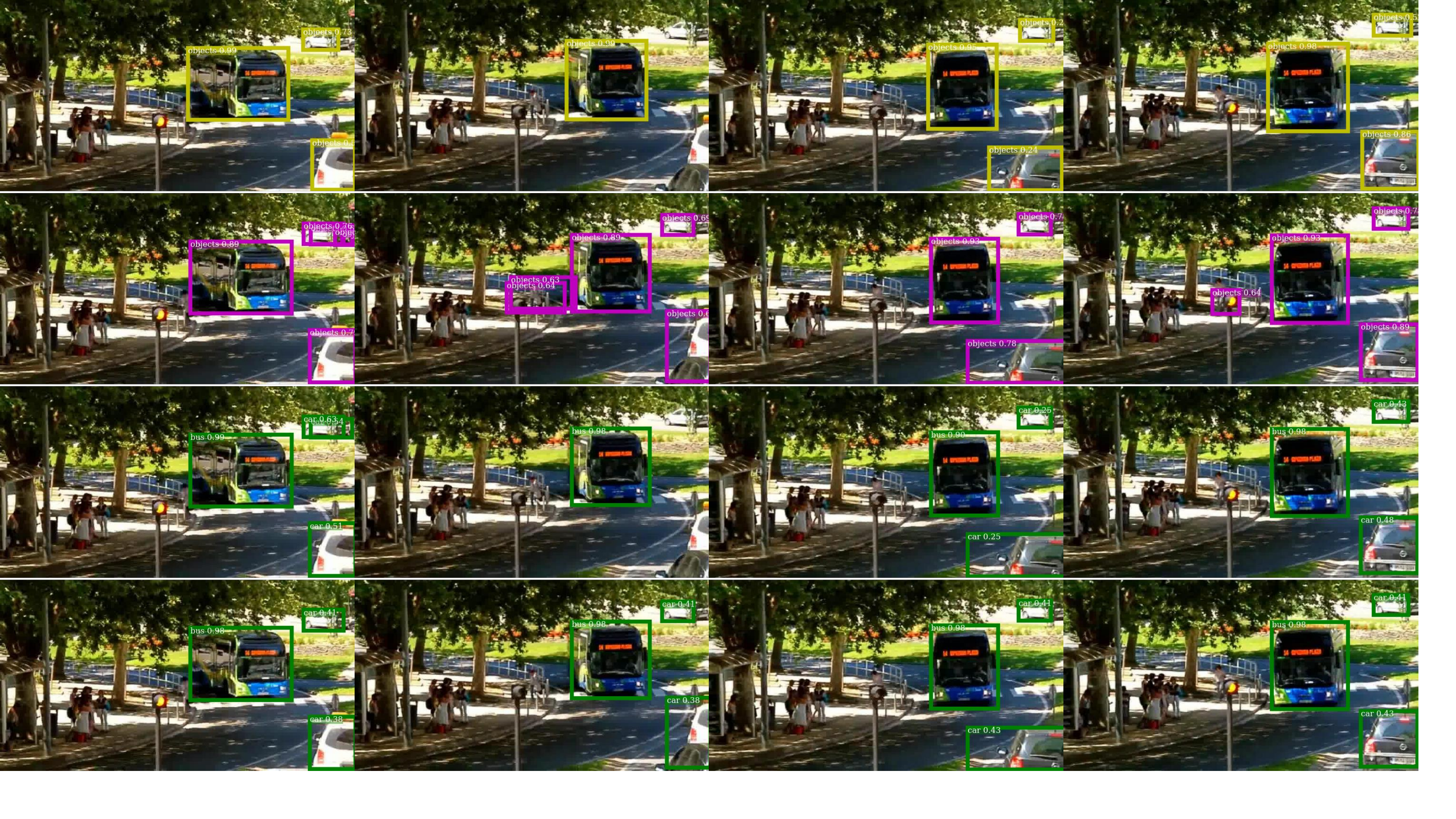}
		\caption{Example of bus and cars. The first row shows the proposed
			object from the detector. The second row shows the proposed object from the tracker. The third row shows the detection from the combined proposals. The fourth row shows the re-scored detection results. Only the objects with scores above 0.2 are shown.}
		\label{fig_3}
	\end{figure*}

	We provide more qualitative video object detection examples in Fig.~\ref{fig_1},Fig.~\ref{fig_2} and Fig.~\ref{fig_3} to show the effectiveness of our method. The tracker may track background objects as well, which is good as their scores for foreground objects will be smoothed to be lower after re-scoring.
	
	\begin{table*}
		\begin{center}
			\begin{adjustbox}{width=1.0\linewidth}
				\begin{tabular}{l|cccccccccccccccc}
					\hline
					Methods &\rotatebox{60}{airplane}&\rotatebox{60}{antelope}&\rotatebox{60}{bear}&\rotatebox{60}{bicycle}&\rotatebox{60}{bird}&\rotatebox{60}{bus}&\rotatebox{60}{car}&\rotatebox{60}{cattle}&\rotatebox{60}{dog}&\rotatebox{60}{d. cat}&\rotatebox{60}{elephant}&\rotatebox{60}{fox}&\rotatebox{60}{g. panda}&\rotatebox{60}{hamster}&\rotatebox{60}{horse}&\rotatebox{60}{lion}\\
					\hline
					HRNet-w32                       & 83.7& 82.8& 79.6& 73.4& 72.0& 68.3& 58.0& 68.5& 65.5& 73.4& 75.8& 86.1& 81.4& 91.9& 69.2& 48.2\\
					HRNet-w32+Seq-NMS               & 83.8& 85.1& 83.8& 73.8& 72.5& 70.0& 59.0& 70.8& 67.9& 78.5& 76.7& 88.7& 81.8& 96.2& 70.6& 58.3\\
					HRNet-w32+TFD(0.3)+Seq-NMS      & 80.9& 84.6& 86.9& 73.6& 74.0& 74.0& 56.4& 82.6& 72.0& 87.8& 76.0& 96.1& 82.3& 98.0& 74.3& 62.3\\
					HRNet-w32+TFD(0.7)+Seq-NMS      & 87.0& 86.0& 85.6& 76.6& 71.7& 75.0& 56.6& 80.8& 72.5& 89.8& 80.4& 95.4& 82.4& 98.8& 79.3& 63.8\\
					ResNeXt101*                     & 87.9& 78.7& 77.5& 73.3& 71.8& 82.5& 60.5& 73.6& 70.7& 82.0& 75.6& 90.7& 86.0& 91.6& 74.4& 57.3\\
					ResNeXt101*+TFD(0.7)+Seq-NMS      & 88.5& 84.3& 81.8& 74.6& 72.5& 89.0& 58.7& 85.2& 79.7& 92.3& 78.6& 98.7& 86.6& 98.8& 80.1& 66.7\\
					ResNeXt101*+TFD(0.7)+Seq-Track-NMS      & 88.3& 84.7& 83.2& 74.6& 72.8& 88.8& 58.5& 85.3& 80.5& 92.3& 78.7& 98.7& 86.2& 98.9& 80.4& 71.2\\
					\hline
					Methods &\rotatebox{60}{lizard}&\rotatebox{60}{monkey}&\rotatebox{60}{motorcycle}&\rotatebox{60}{rabbit}&\rotatebox{60}{red panda}&\rotatebox{60}{sheep}&\rotatebox{60}{snake}&\rotatebox{60}{squirrel}&\rotatebox{60}{tiger}&\rotatebox{60}{train}&\rotatebox{60}{turtle}&\rotatebox{60}{watercraft}&\rotatebox{60}{whale}&\rotatebox{60}{zebra}&\multicolumn{2}{|c}{\rotatebox{60}{mAP (\%)}}\\
					\hline
					HRNet-w32                       & 82.4& 47.1& 81.2& 70.8& 81.2& 55.1& 73.5& 56.2& 89.7& 78.0& 79.2& 63.8& 70.1& 91.0&\multicolumn{2}{|c}{73.2}\\
					HRNet-w32+Seq-NMS               & 84.0& 49.4& 83.3& 75.0& 87.9& 57.3& 74.2& 57.2& 90.2& 78.3& 80.7& 65.2& 72.5& 91.0&\multicolumn{2}{|c}{75.5}\\
					HRNet-w32+TFD(0.3)+Seq-NMS      & 86.8& 48.6& 83.9& 80.5& 94.8& 55.3& 74.9& 63.1& 90.0& 80.3& 82.7& 70.3& 67.6& 93.6&\multicolumn{2}{|c}{77.8}\\
					HRNet-w32+TFD(0.7)+Seq-NMS      & 88.3& 51.7& 87.8& 80.2& 93.2& 58.3& 73.7& 57.3& 89.9& 82.3& 83.0& 72.2& 67.1& 91.0&\multicolumn{2}{|c}{78.6}\\
					ResNeXt101*                     & 79.2& 52.2& 82.2& 74.9& 71.5& 61.8& 79.7& 58.2& 91.9& 86.0& 81.2& 67.4& 73.6& 91.5&\multicolumn{2}{|c}{76.2}\\
					ResNeXt101*+TFD(0.7)+Seq-NMS      & 83.0& 55.1& 87.7& 82.0& 76.4& 64.4& 77.1& 69.2& 91.7& 86.4& 85.2& 70.5& 72.9& 94.0&\multicolumn{2}{|c}{80.4}\\
					ResNeXt101*+TFD(0.7)+Seq-Track-NMS      & 82.8& 55.2& 87.7& 82.1& 90.8& 64.4& 76.9& 69.1& 91.7& 86.4& 85.3& 70.7& 72.4& 94.6&\multicolumn{2}{|c}{81.1}\\
					\hline
				\end{tabular}
			\end{adjustbox}
		\end{center}
		\caption{Ablation study of our method. TFD stands for tracking first detection. * stands for with FPN~\cite{FPN} and DCN~\cite{deformable_conv}.}
		\label{tab_ablation}
	\end{table*}
	
	
\end{document}